\documentclass{article}

\newcommand{\choice}[2]{#2} 

\usepackage{graphicx}
\usepackage[sectionbib]{natbib}
\usepackage{notation}
\choice{
\usepackage[accepted]{icml2012} \usepackage[colorlinks=true,linkcolor=mydarkblue,citecolor=mydarkblue,pdfauthor={Marek Petrik}]{hyperref}}{
\usepackage[colorlinks=true,linkcolor=mydarkblue,citecolor=mydarkblue,pdfauthor={Marek Petrik}]{hyperref}
\usepackage{geometry}
}
\usepackage[articlestyle]{mpemath}
\usepackage[capitalize]{cleveref}

\choice{
\usepackage{amssymb}
\usepackage{amsfonts}}{
\usepackage[bitstream-charter]{mathdesign}
\usepackage[T1]{fontenc}
}

\newcommand{\opt}{{^\star}}
\newcommand{\fmap}{U}

\crefname{prop}{Proposition}{Proposition}
\Crefname{prop}{Proposition}{Proposition}

\crefname{thm}{Theorem}{Theorem}
\Crefname{thm}{Theorem}{Theorem}

\crefname{defn}{Definition}{Definition}
\Crefname{defn}{Definition}{Definition}

\crefname{asm}{Assumption}{Assumption}
\Crefname{asm}{Assumption}{Assumption}

\crefname{lem}{Lemma}{Lemma}
\Crefname{lem}{Lemma}{Lemma}

\choice{\icmltitlerunning{Distributionally Robust Approximate Dynamic Programming}}{
\title{Approximate Dynamic Programming By Minimizing Distributionally Robust Bounds}
\author{Marek Petrik}
\date{}
}

\setitemize{nolistsep}
\setenumerate{nolistsep}
\setdescription{nolistsep}

\newcommand{\stateactions}{\mathcal{W}}
\newcommand{\values}{\mathcal{V}}
\newcommand{\freqs}{\mathcal{U}}

\newcommand{\frep}{\tilde{\fmap}}
\newcommand{\return}{\rho}
\newcommand{\greedy}{\mathcal{G}}
\renewcommand{\cite}{\citep}

\begin{document}

\choice{
\twocolumn[
\icmltitle{Approximate Dynamic Programming By Minimizing Distributionally Robust Bounds}
\vskip 0.3in

\icmlauthor{Marek Petrik}{mpetrik@us.ibm.com}
\icmladdress{IBM T.J. Watson Research Center,
Yorktown, NY, USA}
]
}
{
\maketitle
}

\begin{abstract}
Approximate dynamic programming is a popular method for solving large  Markov decision processes. This paper describes a new class of approximate dynamic programming~(ADP) methods---distributionally robust ADP---that address the curse of dimensionality by minimizing a pessimistic bound on the policy loss. This approach turns ADP into an optimization problem, for which we derive new mathematical program formulations and analyze its properties. DRADP improves on the theoretical guarantees of existing ADP methods---it guarantees convergence and $L_1$ norm-based error bounds. The empirical evaluation of DRADP shows that the theoretical guarantees translate well into good performance on benchmark problems.
\end{abstract}

\section{Introduction} \label{sec:introduction}

Large Markov decision processes~(MDPs) are common in reinforcement learning and operations research and are often solved by approximate dynamic programming~(ADP). Many ADP algorithms have been developed and studied, often with impressive empirical performance. However, because many ADP methods must be carefully tuned to work well and offer insufficient theoretical guarantees, it is important to develop new methods that have both good theoretical guarantees and empirical performance.

Approximate linear programming~(ALP)---an ADP method---has been developed with the goal of achieving convergence and good theoretical guarantees~\citep{Farias2003}. Approximate bilinear programming~(ABP) improves on the theoretical properties of ALP at the cost of additional computational complexity~\citep{Petrik2009e,Petrik2011}. Both ALP and ABP  provide guarantees that rely on conservative error bounds in terms of the $L_\infty$ norm and often under-perform in practice~\citep{Petrik2009e}.
It is, therefore, desirable to develop ADP methods that offer both tighter bounds and better empirical performance.

In this paper, we propose and analyze \emph{distributionally robust approximate dynamic programming} (DRADP)---a new approximate dynamic programming method. DRADP improves on approximate linear and bilinear programming both in terms of  theoretical properties and empirical performance. This method builds on approximate linear and bilinear programming but achieves better solution quality by explicitly optimizing tighter, less conservative, bounds stated in terms of a weighted $L_1$ norm. In particular, DRADP computes a good solution for a given initial distribution instead of attempting to find a solution that is  good for all initial distributions.

The objective in ADP is to compute a policy $\pol$  with the maximal return $\return(\pol)$. Maximizing the return also minimizes the loss with respect to the optimal policy $\pol\opt$---known as the \emph{policy loss} and defined as $\return(\pol\opt) - \return(\pol)$. There are two main challenges in computing a good policy for a large MDP. First, it is necessary to efficiently evaluate its return; evaluation using simulation is time consuming and often impractical. Second, the return of a parameterized policy may be a function that is hard to optimize. DRADP addressed both these issues by maximizing a simple \emph{lower bound} $\tilde\return(\pol) $ on the return using ideas from robust optimization. This lower bound is easy to optimize and can be computed from a  small sample of the domain, eliminating the need for extensive simulation.

Maximizing a lower bound on the return corresponds to minimizing an upper bound $\return(\pol\opt) - \tilde\return(\pol)$ on the policy loss. The main reason to minimize an upper bound---as opposed to a lower bound---is that the approximation error can be bounded by the difference $\return(\pol\opt) - \tilde\return(\pol\opt)$ for the optimal policy only, instead of the difference for the set of all policies, as we show formally in \cref{sec:error}.

The lower bound  on the return in DRADP is based on an approximation of the \emph{state occupancy distributions} or frequencies~\citep{Puterman2005}. The state occupancy frequency represents the fraction of time that is spent in the state and is in some sense the dual of a value function. Occupancy frequencies have been used, for example, to solve factored MDPs~\citep{Dolgov2006} and in dual dynamic programming~\citep{Wang2007,Wang2008} (The term ``dual dynamic programming'' also refers to unrelated linear stochastic programming methods). These methods can improve  the empirical performance, but proving bounds on the policy loss has proved challenging. We take a different approach to prove tight bounds on the policy loss. While the existing methods approximate the state occupancy frequencies by a \emph{subset}, we  approximate it by a \emph{superset}.

We call the DRADP approach distributionally robust because it uses the robust optimization methodology to represent and simplify the set of occupancy distributions~\citep{Delage2010}. \emph{Robust optimization} is a recently revived approach for modeling uncertainty in optimization problems~\cite{Ben-Tal2009}. It does not attempt to model the uncertainty precisely, but instead computes solutions that are immunized against its effects. In distributionally robust optimization, the uncertainty is in probability distributions. DRADP introduces the uncertainty in state occupancy frequencies in order to make very large MDPs tractable and uses the robust optimization approach to compute solutions that are immune to this uncertainty.

The remainder of the paper is organized as follows. First, in \cref{sec:framework}, we define the basic framework including MDPs and value functions. Then, \cref{sec:dradp} introduces the general DRADP method in terms of generic optimization problems. \cref{sec:error} analyzes approximation errors involved in DRADP and shows that standard concentration coefficient assumptions on the MDP~\cite{Munos2007} can be used to derive tighter bounds. To leverage existing mathematical programming  methods, we show that DRADP can be formulated in terms of standard mathematical optimization models in \cref{sec:computational}. Finally, \cref{sec:experimental} presents experimental results on standard benchmark problems. \choice{Due to space constraints we omit the proofs in this version; please see the extended version for the full proofs~\cite{Petrik2012ext}.}{}

We consider the offline---or batch---setting in this paper, in which  all samples are generated in advance of computing the value function. This setting is identical to that of LSPI~\cite{Lagoudakis2003} and ALP~\cite{Farias2003}.

\section{Framework and Notation} \label{sec:framework}

In this section, we define the basic concepts required for solving Markov decision processes: value functions, and  occupancy frequencies. We use the following general notation throughout the paper. The symbols $\zero$ and $\one$ denote vectors of all zeros or ones of appropriate dimensions respectively; the symbol $\eye$ denotes an identity matrix of an appropriate dimension. The operator $\pos{\cdot}$ denotes an element-wise non-negative part of a vector. We will often use linear algebra and expectation notations interchangeably; for example: $\Ex{X}{u} = u\tr x$, where $x$ is a vector of the values of the random variable $X$. We also use $\Real^{\mathcal{X}}$ to denote the set of all functions from a finite set $\mathcal{X}$ to $\Real$; note that $\Real^{\mathcal{X}}$ is trivially a vector space.

A \emph{Markov Decision Process} is a tuple $(\states,\actions,\tran,\rew,\indist)$. Here, $\states$ is a  \emph{finite} set of states, $\actions$ is a \emph{finite} set of actions, $\tran: \states \times \actions \times \states \mapsto [0,1]$ is the transition function ($\tran(s,a,s')$ is the probability of transiting to state $s'$ from state $s$ given action $a$), and $\rew: \states \times \actions \mapsto \Real$ is a reward function. The initial distribution is: $\indist: \states \mapsto [0,1]$, such that $\sum_{s\in\states} \indist(s) = 1$. The set of all state-action pairs is $\stateactions = \states\times\actions$. For the sake of simplicity, we assume that all actions can be taken in all states. To avoid technicalities that detract from the main ideas of the paper, we  assume finite state and action sets but the results  apply with additional compactness assumptions to infinite sets. We will use $S$ and $W$ to denote random variables with values in $\states$ and $\stateactions$.

The solution of an MDP is a stationary \emph{deterministic policy} $\pol:\states\rightarrow\actions$, which determines the action to take in any state; the set of all deterministic policies is denoted by $\policies_D$. A stationary \emph{randomized---or stochastic---policy} $\pol: \states \times \actions \rightarrow [0,1]$ assigns the probability to all actions for every state; the set of all randomized policies is denoted by $\policies_R$. Clearly $\policies_D \subseteq \policies_R$ holds by mapping the chosen action to the appropriate distribution. A randomized policy can be thought of as a vector on $\stateactions$ that assigns the appropriate probabilities to each state--action pair.

For any  $\pol\in\policies_R$, we can define the transition probability matrix and the reward vector as follows: $\tran_\pol(s,s') = \sum_{a \in \actions} \tran(s,a,s')\cdot\pol(s,a)$ and $\rew_\pol(s) = \sum_{a \in \actions} \rew(s,a)\cdot\pol(s,a)$. We use $P_a$ and $r_a$ to represent values for a policy that always takes action $a\in\actions$. We also define a matrix $\tmat$ and a vector $\trew$ as follows:
\begin{align*}
\tmat\tr &= \begin{pmatrix} \eye - \disc \tran_{a_1}\tr & \eye - \disc \tran_{a_2}\tr & \cdots \end{pmatrix},  & \trew\tr &= \begin{pmatrix} \rew_{a_1}\tr & \rew_{a_2}\tr & \cdots \end{pmatrix}
\end{align*}

Values $\tmat$ and $\trew$ are usually used in approximate linear programming~(ALP)~\cite{Schweitzer1985} and linear program formulations of MDPs~\cite{Puterman2005}. The main objective in solving an MDP is to compute a policy with the maximal return.
\begin{defn}
The return $\return:\policies_R\rightarrow\Real$ of $\pol\in\policies_R$ is defined as:
$\return(\pol) = \sum_{n=0}^{\infty} \indist\tr (\disc \cdot \tran_\pol)^n \; \rew_\pol$.
The \emph{optimal policy} solves  $\pol\opt \in \arg\max_{\pol\in\policies_R} \return(\pol)$ and we use $\return\opt = \return(\pol\opt)$.
\end{defn}

DRADP relies on two main solution concepts: \emph{state occupancy frequencies} and \emph{value functions}. State occupancy frequencies---or measures---intuitively represent the probability of terminating in each state when the discount factor $\disc$ is interpreted as a probability of remaining in the system~\cite{Puterman2005}. State-action occupancy frequencies are defined  for state--action pairs and represent the joint probability of being in the state and taking the action.

\emph{State occupancy frequency} for $\pol\in\policies_R$ is denoted by $d_\pol \in \Real^{\states}$ and is defined as:
\[ d_\pol = (1 - \disc)\cdot \sum_{t=0}^\infty (\disc \cdot\tran_\pol\tr)^t \indist = (1 - \disc)\cdot \invm{\eye - \disc\cdot\tran_\pol\tr} \indist~.\]
\emph{State-action occupancy frequency} is denoted by $u_\pol \in  \Real^{\stateactions}$(its set-valued equivalent is $\fmap(\pol)$) and is a product of state--occupancy frequencies and action probabilities:
\[ u_\pol(s,a) = d_\pol(s)\cdot \pol(s,a)~,  \quad  \fmap(\pol) = \{ u_\pol \}~. \]
Note that $\fmap(\pol)$ is a set-valued function with the output set of cardinality 1. State and state-action occupancy frequencies represent valid probability measures over $\states$ and $\stateactions$ respectively. We use $d\opt = d_{\pol\opt}$ and $u\opt = u_{\pol\opt}$ to denote the optimal measures. Finally, we use $u\vert_\pol \in \RealPlus^\states$ for $\pol\in\policies_D$ to denote a restriction of $u$ to $\pol$ such that $u\vert_\pol (s) = u(s,\pol(s))$.

State-action occupancy frequencies are closely related to the set $\freqs$ of dual feasible solutions of the linear program formulation of an MDP, which is defined as~(e.g. Section 6.9 of \cite{Puterman2005}):
\begin{equation} \label{eq:frequencies}
\freqs = \left\{ u \in \RealPlus^{\stateactions} \setsep \tmat\tr u = (1-\disc) \cdot\indist \right\}.
\end{equation}
The following well-known proposition characterizes the basic properties of the set $\freqs$.
\begin{prop}[e.g. Theorem 6.9 in \cite{Puterman2005}]\label{prop:distrib_set}
The set of occupancy frequencies satisfies the following properties.
\begin{enumerate}
\item[(i)] $\freqs = \bigcup_{\pol\in\policies_R} \fmap(\pol) = \operatorname{conv}(\bigcup_{\pol\in\policies_D} \fmap(\pol))$.
\item[(ii)] For each $\bar u\in\freqs$, define $\pol'(s,a) = \bar{u}(s,a) /\sum_{a'\in\actions} \bar{u}(s,a')$. Then $u_{\pol'} = \bar u$.
\item[(iii)] $\one\tr u = 1$ for each $u \in \freqs$.
\item[(iv)] $\tmat\tr u = (1-\disc) \cdot \alpha$ for each $u \in \freqs$.
\end{enumerate}
\end{prop}
Part (i), in particular, holds because deterministic policies represent the basic feasible solutions of the dual linear program for an MDP.

A \emph{value function} $\val_\pol \in  \Real^{\states}$ of $\pol\in\policies_R$ maps states to the return obtained when starting in them and is defined by:
\[ \val_\pol = \sum_{t=0}^\infty (\disc \cdot\tran_\pol)^t \rew_\pol = \invm{\eye - \disc \cdot\tran_\pol} \rew_\pol~. \]
The set of all possible value functions is denoted by $\values$. It is well known that a policy $\pol\opt$ with the value function $v\opt$ is \emph{optimal} if and only if   $\val\opt \ge \val_\pol$ for every $\pol\in\policies_R$. The value function update $\Bell_\pol$ for a policy $\pol$ and the Bellman operator $\Bell$ are defined as: $\Bell_\pol v = \disc\tran_\pol v + r_\pol$ and $\Bell v = \max_{\pol\in\policies_R} \Bell_\pol \val$.

The optimal value function $\val\opt$ satisfies $\Bell \val\opt = \val\opt$.
The following proposition states the well-known connection between state--action occupancy frequencies and value functions.
\begin{prop}[e.g. Chapter 6 in \cite{Puterman2005}] \label{prop:return_frequencies}
For each $\pol\in\policies_R$: $\return(\pol) = \Ex{\val_\pol(S)}{\indist} = \Ex{\rew(W)}{u_\pol} / (1-\disc)~.$
\end{prop}

The value function, computed by a dynamic programming algorithm, is typically then used to  derive the greedy policy. A greedy policy takes in every state an action that maximizes the expected conditional return.
\begin{defn} \label{def:greedy_policy}
A policy $\pol \in \policies_D$ is \emph{greedy} with respect to a value function $\val$ when $\Bell_\pol \val = \Bell \val$; in other words:
\[ \pol(s) \in \arg\max_{a\in\actions} ~\Bigl( ~\rew(s,a) + \disc \cdot \sum_{s'\in\states} \tran(s,a,s')\cdot \val(s')~ \Bigr), \]
for each $s\in\states$ with ties broken arbitrarily.
\end{defn}

MDP is a very general model. Often, specific properties of the MDP can be used to compute better solutions and to derive tighter bounds. One common assumption---used to derive $L_2$ bounds for API---is a smoothness of transition probabilities~\cite{Munos2003}, also known as the concentration coefficient~\cite{Munos2007}; this property can be used to derive tighter DRADP bounds.
\begin{asm}[Concentration coefficient] \label{asm:smooth}
There exists a probability measure $\mu \in [0,1]^\states$ and a constant $C\in\RealPlus$ such that for all $s,s'\in\states$ and all $\pol\in\policies_D$ the transition probability is bounded as:
$\tran(s,\pol(s),s') \leq C \cdot \mu(s').$
\end{asm}

\section{Distributionally Robust Approximate Dynamic Programming} \label{sec:dradp}

In this section, we formalize DRADP and describe it in terms of generic optimization problems. Practical DRADP implementations are sampled versions of the optimization problems described in this section. However, as it is common in ADP literature, we do not explicitly analyze the sampling method used with DRADP in this paper, because the sampling error can simply be added to the error bounds that we derive. The sampling is performed and errors bounded identically to approximate linear programming and approximate bilinear programming---state and action samples are used to select a subset of constraints and variables~\cite{Farias2003,Petrik2010,Petrik2011}.

The main objective of ADP is to compute a policy $\pol\in\policies_R$ that maximizes the return $\return(\pol)$. Because the MDPs of interest are very large, a common approach is to simplify them by restricting the set of policies that are considered to a smaller set $\tilde\policies \subseteq \policies_R$. For example, policies may be constrained to take the same action in some states; or  to be greedy with respect to an approximate value function. Since it is not possible to compute an optimal policy, the common objective is to minimize the policy loss. Policy loss captures the difference in the discounted  return when following policy $\pol$ instead of the optimal policy $\pol\opt$.
\begin{defn}\label{def:policy_loss}
The \emph{expected policy loss} of $\pol\in\policies_R$ is defined as:
\[\return\opt - \return(\pol) =  \frac{\trew\tr (u\opt - u_\pol)}{1-\disc} = \| \val\opt - \val_\pol\|_{1,\indist}~,\]
where $\|\cdot\|_{1,\indist}$ represents an $\indist$-weighted $L_1$ norm.
\end{defn}

ADP  relies on approximate value functions $\rep \subseteq \values$ that are a \emph{subset} of all value functions. In DRADP, approximate value functions are used simultaneously  to both restrict the space of policies and to approximate their returns. We, in addition, define a set of approximate occupancy frequencies $\frep(\pol) \supseteq \fmap(\pol)$ that are a \emph{superset} of the true occupancy frequencies. We call any  element in the appropriate approximate sets  \emph{representable}.

We consider \emph{linear function approximation}, in which the values for states are represented as a linear combination of \emph{nonlinear basis functions (vectors)}. For each $s\in\states$, we define a vector $\phi(s)$ of features with $|\phi|$ being the dimension of the vector. The rows of the basis matrix $\repm$ correspond to $\phi(s)$, and the approximation space is generated by the columns of $\repm$. Approximate value functions and \emph{policy-dependent} state occupancy measures  for linear approximations are defined for some given feature matrices $\repm_u$ and $\repm_v$ as:
\begin{align}
\label{eq:rep_values} \rep &= \Bigl\{ ~\val \in \values \setsep \val = \repm_v x, \; x \in \Real^{|\phi|} ~\Bigr\}~, \\
\label{eq:rep_dists} \frep(\pol) &= \Bigl\{ ~u \in \RealPlus^{\actions} \setsep
\begin{matrix}
\repm_u\tr \tmat\tr u = (1-\disc)\cdot \repm_u\tr\indist, \\
u(s,a) \le \pol(s,a)
\end{matrix} \Bigr\} ~.
\end{align}
Clearly, $\frep(\pol) \supseteq \fmap(\pol)$ from the definition of $u_\pol$.  We will assume the following important assumption without reference for the remainder of the paper.
\begin{asm}\label{asm:contains_one}
One of the features in each of $\feats_u$ and $\feats_v$ is a constant; that is, $\one = \repm_u x_u $ and $\one= \repm_v x_v$ for some $x_u$ and $x_v$.
\end{asm}
The following lemma, which can be derived directly from the definition of $\frep$ and \cref{prop:distrib_set}, shows the importance of \cref{asm:contains_one}.
\begin{lem} \label{lem:fixed_sum}
Suppose that \cref{asm:contains_one} holds. Then for each $\pol\in\policies_R$:
$u \in \frep(\pol) \Rightarrow \one\tr u = 1~.$
\end{lem}

Approximate policies $\tilde\policies$ are most often represented indirectly---by assuming policies that are greedy to the approximate value functions. The set $\greedy$ of all such greedy policies is defined by:
$\greedy = \{ ~\pol\in\policies_D \setsep \Bell_\pol \val = \Bell \val, \val \in \rep ~\}$.
Although DRADP applies to other approximate policy sets we will particularly focus on the set $\tilde\policies = \greedy$.

We are now ready to define the basic DRADP formulation which is analyzed in the remainder of the paper.
\begin{defn} \label{def:dradp}
DRADP computes an approximate policy by solving the following optimization problem:
\begin{equation} \label{eq:main_optimization} \tag{DRADP}
\arg \max_{\pol\in\tilde\policies} \tilde\return(\pol) = \arg \min_{\pol\in\tilde\policies} ~\bigl(~ \return\opt - \tilde\return(\pol) ~\bigr)~   ,
\end{equation}
where the function $\tilde\return : \policies_R\rightarrow \Real$ is defined by:
\begin{equation} \label{eq:bound_both}
\tilde\return(\pol) = \max_{\val\in\rep} ~ \Bigl(~ \indist\tr \val - \max_{u \in \frep(\pol)}  \frac{u\tr (\tmat \val - \trew)}{1-\disc} ~\Bigr)  ~.
\end{equation}
\end{defn}
Note that the solution of \eqref{eq:main_optimization} is a policy; this policy is not necessarily greedy with respect to the optimal $\val$ in \eqref{eq:bound_both} unlike in most other ADP approaches. The expression \eqref{eq:bound_both} can be understood intuitively as follows. The first term, $\indist\tr v$, represents the expected return if $\val$ is the value function of $\pol$. The second term $\max_{u \in \frep(\pol)}  (u\tr (\tmat \val - \trew))/(1-\disc)$ is a penalty function, which offsets any gains when $v \neq v_\pol$ and is motivated by the primal-dual slack variables in the LP formulation of the MDP. Given this interpretation, DRADP simultaneously restricts the set of value functions and upper-approximates the penalty function.

The following theorem states an important property of \cref{def:dradp}, which is used to derive approximation error bounds.
\begin{thm}\label{thm:mainloss}
For each $\pol\in\policies_R$, $\tilde\return$  lower-bounds the true return: $\tilde\return(\pol) \le \return(\pol)$.
In addition, when $\repm_u$ and $\repm_v$ are invertible and $\pol\in\policies_D$ then $\return(\pol) = \tilde\return(\pol)$.
\end{thm}
\choice{}{
\begin{proof}[Proof of \cref{thm:mainloss}]
The theorem follows by complementary slackness from the fact that $\frep(\pol) \supseteq \fmap(\pol)$.
We use that $d_\pol\tr~ (\eye - \disc \tran_\pol) = (1-\disc) \cdot \indist\tr$ for any $\pol\in\policies_R$ from the definition of $d_\pol$ to derive:
\begin{align*}
\return(\pol) &= \Ex{\val_\pol}{\indist} = \indist\tr \val_\pol
= \frac{\rew_\pol\tr d_\pol}{1-\disc} + \max_{\val\in\rep} \left(\indist\tr - \frac{d_\pol\tr (\eye-\disc\cdot\tran_\pol)}{1-\disc} \right) \val \\
&= \frac{\rew_\pol\tr d_\pol}{1-\disc} + \max_{\val\in\rep} ~\Bigl(~ \indist\tr \val - \frac{d_\pol\tr (\eye-\disc\cdot\tran_\pol) \val}{1-\disc}  ~\Bigr) \\
&=\max_{\val\in\rep}  ~\Bigl(~ \indist\tr \val - \frac{d_\pol\tr (\eye-\disc\cdot\tran_\pol)\val - d_\pol\tr \rew_\pol}{1-\disc}~\Bigr) \\
&= \max_{\val\in\rep} ~\Bigl(~ \indist\tr \val - \frac{d_\pol\tr\left( \val - \Bell_\pol \val \right)}{1-\disc}\Bigr) \\
&= \max_{\val\in\rep} ~\Bigl(\indist\tr \val-~ \frac{u_\pol\tr\left( A\val - b \right)}{1-\disc}   ~\Bigr) \\
&\stackrel{\text{\eqref{eq:rep_dists}}}{\ge} \max_{\val\in\rep} \min_{u\in\frep(\pol)} ~\Bigl(~\indist\tr \val - \frac{u\tr\left( A\val - b \right)}{1-\disc} ~\Bigr) \\
&= \tilde\return(\pol)~.
\end{align*}
Now, assume that $\repm_u$ is the identity matrix. Then, using (i) of \cref{prop:distrib_set}, $\frep(\pol) = \{ u_\pol \}$ whenever $\pol\in\policies_D$. Note that this does not hold for $\pol\in\policies_R\setminus\policies_D$.
\end{proof}}

We now show that the lower bound $\tilde\return$ in \eqref{eq:bound_both} can be simplified in some cases by ignoring the value functions for any $\pol\in\policies_R$; the formulation \eqref{eq:bound_both} will nevertheless be particularly useful in the theoretical analysis because it relates value functions and occupancy frequencies.
\begin{equation} \label{eq:main_simple}
\tilde\return'(\pol) = \min_{u \in \frep'(\pol) } \frac{u\tr \trew}{1-\disc} ~,
\end{equation}
where $\frep'(\pol)$ is defined equivalently to $\frep(\pol)$ with the exception that $\repm_u = \repm_v = \repm$ for some $\repm$.
\begin{prop}\label{prop:return_simple}
When $\repm_v = \repm_u$, then $\tilde\return(\pol) = \tilde\return'(\pol)$. When $\repm_v \neq \repm_u$, then define $\frep'$ and $\tilde\return'$ using a new representation $\repm' = [\repm_v  ~\repm_u]$. Then: $\tilde\return(\pol) = \tilde\return'(\pol).$
\end{prop}
\choice{}{
\begin{proof}[Proof of \cref{prop:return_simple}]
Recall that:
\[\frep(\pol) = \left\{ u \in \RealPlus^{\actions} \setsep \repm_u\tr \tmat\tr u = (1-\disc)\cdot \repm_u\tr\indist,  ~ u(s,a) \le \pol(s,a) \right\} ~.\]
Assume that $\repm = \repm_v = \repm_u$. First, note that $u\in\frep(\pol)$  implies that $\repm_u\tr \tmat u =(1-\disc) \repm_u\tr \indist$ and $\val\in\rep$ implies that $\val = \repm_v x$ for some $x$. Then:
\begin{align*}
\tilde\return(\pol)  &= \max_{\val\in\rep} \min_{u \in \frep(\pol)} ~\Bigl(~ \indist\tr \val - \frac{u\tr (\tmat \val - \trew)}{1-\disc} ~\Bigr)  \\ &\stackrel{\text{\eqref{eq:rep_values}}}{=} \max_x \min_{u \in \frep(\pol)} ~\Bigl(~\indist\tr \repm x - \frac{u\tr (\tmat \repm x - \trew)}{1-\disc}  ~\Bigr) \\
&\stackrel{\text{\eqref{eq:rep_dists}}}{=} \max_x \min_{u \in \frep(\pol)} ~\Bigl(~\indist\tr \repm x - \frac{(1-\disc) \indist\tr \repm x}{1-\disc}  + \frac{u\tr \trew}{1-\disc} ~\Bigr) \\
&= \max_x \min_{u \in \frep(\pol)}~ \frac{u\tr b }{1-\disc} \\
&= \min_{u \in \frep(\pol)} ~\frac{u\tr \trew}{1-\disc} ~.
\end{align*}
When $\repm_v \neq \repm_u$, define $\repm = [\repm_v ~\repm_u]$. Then:
\begin{align*}
\tilde\return(\pol) &\stackrel{\text{\eqref{eq:rep_values}}}{=} \max_x \min_{u \in \frep(\pol)} ~\Bigl(~ \indist\tr \repm_v x - \frac{u\tr (\tmat \repm_v~ x - \trew)}{1-\disc} ~\Bigr) \\
&= \max_x ~\Bigl(~\indist\tr  \repm _v x - \max_{u \in \frep(\pol)}   \frac{u\tr (\tmat \repm_v~ x - \trew)}{1-\disc} ~\Bigr) \\
&\stackrel{(\star)}{=} \max ~\Bigl\{~ \indist\tr\bigl((1-\disc)\cdot  \repm_u \lambda_1 + \repm_v x \bigr) - \pol\tr \lambda_2 \setsep  \tmat \Bigl( \repm_u \lambda_1 +\frac{ \repm_v ~x}{1-\disc} \Bigr) - b \le \lambda_2, ~\lambda_2 \ge \zero ~\Bigr\} \\
&= \max~ \Bigl\{~ \indist\tr [\repm_u ~\repm_v]\tr y  - \pol\tr \lambda_2 \setsep  \tmat [\repm_u  ~\repm_v]\tr y  -b\le  (1-\disc)\cdot \lambda_2, \lambda_2 \ge \zero ~\Bigr\} \\
&\stackrel{(\star)}{=} \min_{u\in \frep_c(\pol) } \frac{u\tr \trew}{1-\disc}~,
\end{align*}
where:
\[ \frep_c(\pol) = \bigl\{ u \setsep [\repm_u ~\repm_v]\tr \tmat\tr u = (1-\disc)\cdot   [\repm_u ~\repm_v]\tr \indist, ~\zero\le u \le \pol  \bigr\}~. \]
Above, $(\star)$ equalities are derived from LP duality. This shows that simply combining the features to  $[\repm_u ~\repm_v]$ leads to the same approximate return function as using them separately.
\end{proof}
}

For the remainder of the paper assume that $\repm_v = \repm_u$ since assuming that they are the same does not reduce the solution quality.

A potential challenge with DRADP is in representing the set of approximate policies $\tilde\policies$, because a policy must generalize to all states even when computed from a small sample. Note, that for a fixed value function $\val$ in \eqref{eq:bound_both} the policy that solves $\min_{\pol\in\tilde\policies} \tilde\return(\pol)$ is not necessarily the greedy with respect to $v$. The following representation theorem, however, shows that when the set of representable policies $\tilde\policies$ is sufficiently rich, then the computed policy will be greedy with respect to a representable value function.
\begin{thm} \label{thm:representation}
Suppose that $\tilde\policies \supseteq \greedy$. Then:
\begin{itemize}
\item[(i)] $\max_{\pol\in\tilde\policies}  \tilde\return(\pol) = \max_{\pol\in\greedy} \tilde\return(\pol)$.
\item[(ii)] $\exists \bar\pol\in\arg\max_{\pol\in\tilde\policies} \tilde\return(\pol)$ such that $\bar\pol\in\greedy$.
\end{itemize}
\end{thm}
\choice{}{
\begin{proof}[Proof of \cref{thm:representation}]
Recall, that given \cref{prop:return_simple}, we can assume the simple representation of the return $\return'(\pol)$. For a fixed $\pol\in\policies_R$ the optimization in $\return'(\pol)$ over $u$ represents the following linear program:
\begin{mprog*}
\minimize{u} u\tr b /(1-\disc)
\stc \repm\tr \tmat\tr u = (1-\disc)\cdot  \repm\tr\indist
\cs \zero\le u \le \pol~.
\end{mprog*}
The dual of this linear program is:
\begin{mprog} \label{eq:dual_dradp_proof}
\maximize{\lambda_1, \lambda_2} \indist\tr \repm \lambda_1  -\pol\tr \lambda_2
\stc \tmat\repm\lambda_1 \le \trew + (1-\disc)\cdot   \lambda_2
\cs \lambda_2 \ge \zero~.
\end{mprog}
Now, suppose a fixed $\lambda_1$ in \eqref{eq:dual_dradp_proof} and let $\val = \repm \lambda_1$. Then, $\pol\ge\zero$ implies there exists one optimal $\lambda_2$ such that:
\[ \lambda_2(s,a) = \pos{v(s) - \disc\cdot   \sum_{s'\in\states}  \tran(s,a,s') \val(s') - \rew(s,a) } \quad \forall s \in\states, ~a\in\actions~. \]
This means that \eqref{eq:dual_dradp_proof} is equivalent to:
\begin{mprog*}
\maximize{\lambda_1} ~\Bigl( \indist\tr \repm \lambda_1  - \frac{\pol\tr \pos{\tmat\repm\lambda_1 - \trew}}{1-\disc} ~\Bigr)
\end{mprog*}
Now, consider policies $\pol_1, \pol_2 \in \policies_R$ that are identical in all states except in $s \in\states$. That is $\pol_1(s) = a_1,\pol_2(s) = a_2$ and assume that $a_1$ is the greedy action. To simplify the notation, we are assuming that $a_2$ can represent a randomized action. Then:
\begin{align*}
\disc\cdot  \sum_{s'\in\states} \tran(s,a_1,s')\cdot   \val(s') - \rew(s,a_1) &\stackrel{\text{D \ref{def:greedy_policy}}}{\ge} \disc\cdot   \sum_{s'\in\states} \tran(s,a_2,s')\cdot   \val(s') - \rew(s,a_2)~, \\
v(s) - \disc\cdot   \sum_{s'\in\states} \tran(s,a_1,s')\cdot   \val(s') - \rew(s,a_1) &\le v(s) - \disc\cdot   \sum_{s'\in\states} \tran(s,a_2,s')\cdot   \val(s') - \rew(s,a_2)~, \\
\pos{v(s) - \disc\cdot   \sum_{s'\in\states} \tran(s,a_1,s')\cdot   \val(s') - \rew(s,a_1)} &\le \pos{ v(s) - \disc\cdot   \sum_{s'\in\states} \tran(s,a_2,s')\cdot   \val(s') - \rew(s,a_2)}~, \\
\pol_1\tr \pos{\tmat\repm\lambda_1 - \trew} &\le \pol_2\tr \pos{\tmat\repm\lambda_1 - \trew}~, \\
\tilde\return'(\pol_1) &\ge \tilde\return'(\pol_2)~.
\end{align*}
Applying this argument inductively to every state, one can construct a policy $\pol\in\greedy \cap \tilde\policies$ with respect to the approximate value function. Because this argument applies to each state individually, it also applies to DRADP from an incomplete sample.
\end{proof}
}

Note that the assumption $\tilde\policies \supseteq \greedy$ simply implies that DRADP can select a policy that is greedy with respect to any \emph{approximate} value function $\val\in\rep$. This is an implicit assumption in many ADP algorithms, including ALP and LSPI. We state the assumption explicitly to indicate results that do not hold in case there are additional restrictions on the set of policies that is considered.

\cref{thm:representation} implies that it is only necessary to consider policies that are greedy with respect to representable value functions which is the most common approach in ADP. However, other approaches for representing policies may have better theoretical or empirical properties and should be also studied.

\section{Approximation Error Bounds} \label{sec:error}

This section describes the \emph{a priori} approximation properties of DRADP solutions; these bounds can be evaluated before a solution is computed. We focus on several types of bounds that not only show the performance of the method, but also make it easier to theoretically compare DRADP to existing ADP methods. These bounds show that DRADP has stronger theoretical guarantees than most other ADP methods. The first bound mirrors some  simple bounds for approximate policy iteration~(API) in terms of the $L_\infty$ norm~\cite{Munos2007}:
\begin{equation}\label{eq:api_bound}
\lim\sup_{k\rightarrow\infty} \| v\opt - \val_{\pol_k} \|_{\infty} \leq \frac{2\cdot\disc}{(1-\disc)^2} \limsup_{k\rightarrow\infty} \epsilon_k~,
\end{equation}
where $\pol_k$ and $\epsilon_k$ are the policy and $L_\infty$ approximation error at iteration $k$.
\begin{thm}\label{thm:simplelossbounds}
Suppose that $\tilde\policies \supseteq \greedy$ and  that $\bar\pol \in\arg\max_{\pol\in\tilde\policies} \tilde\return(\pol)$ in \eqref{eq:main_optimization}. The policy loss $\return\opt - \return(\bar\pol)$ is then bounded as:
\begin{equation} \label{eq:dradp_bound_simple}
\|\val\opt - \val_{\bar\pol}\|_{1,\indist} \le \frac{ 2}{1-\disc} \min_{\val\in\rep}  \|  \val-\Bell\val \|_\infty~.
\end{equation}
\end{thm}
\choice{}{
\begin{proof}[Proof of \cref{thm:simplelossbounds}]
Assume that $\bar \pol$ is an optimal solution of \eqref{eq:main_optimization}. First, the policy loss can  be expressed in terms of approximate values and occupancy frequencies:
\begin{align*}
\return\opt - \return(\bar\pol) &\stackrel{\text{T \ref{thm:mainloss}}}{\le} \return\opt - \tilde\return(\bar\pol) = \min_{\pol\in\tilde\policies} ~(~\return\opt - \tilde\return(\pol)~) \stackrel{\greedy\subseteq\tilde\policies}{\le}\min_{\pol\in\greedy}  ~(~ \return\opt - \tilde\return(\pol)~) \\
&= \min_{v\in\rep} \min_{\pol\in\greedy}  \max_{u\in\frep(\pol)} ~\Bigl(~ \frac{u\tr (\tmat v - \trew) }{1-\disc} + \indist\tr (v\opt - v ) ~\Bigr) \\
&\stackrel{\text{P \ref{prop:distrib_set}}}{=} \frac{1}{1-\disc} \min_{v\in\rep} \min_{\pol\in\greedy}  \max_{u\in\frep(\pol)} ~\Bigl(~ u\tr (\tmat v - \trew) + (u\opt)\tr\tmat (v\opt - v )  ~\Bigr) \\
&\stackrel{\text{P \ref{prop:return_frequencies}}}{=} \frac{1}{1-\disc} \min_{v\in\rep} \min_{\pol\in\greedy}  \max_{u\in\frep(\pol)} ~\Bigl(~ u\tr (\tmat v - \trew) + (u\opt)\tr ( \trew - \tmat v ) ~\Bigr) \\
&\stackrel{\text{\eqref{eq:rep_dists}}}{\le} \frac{1}{1-\disc} \min_{v\in\rep} \min_{\pol\in\greedy}  \max_{u_1,u_2\in\frep(\pol)} ~\Bigl(~ u_1\tr (\tmat v - \trew) + u_2\tr ( \trew - \tmat v) ~\Bigr) \\
&\stackrel{u_i \ge \zero}{\le} \frac{1}{1-\disc} \min_{v\in\rep} \max_{u_1,u_2\in\frep(\pol)} ~\Bigl(~ u_1\tr \pos{\tmat\val - \trew} + u_2\tr \pos{\trew - \tmat\val} ~\Bigr) \\
&\le \frac{2}{1-\disc} \min_{v\in\rep} \max_{u\in\frep(\pol)} ~\Bigl(~ u\tr |\tmat\val - \trew| ~\Bigr)~.
\end{align*}
Then, the loss can be upper bounded by choosing the greedy policy:
\begin{align*}
\return\opt - \return(\bar\pol) &\le \frac{2}{1-\disc} \min_{v\in\rep} \min_{\pol\in\greedy} \max_{u\in\frep(\pol)} ~\Bigl(~ u\tr |\tmat \val - \trew| ~\Bigr) \\
&\le \frac{2}{1-\disc} \min_{v\in\rep} \min_{\pol\in\greedy} \max_{u\in\frep(\pol)} ~\Bigl(~ u\vert_\pol\tr |\val - \Bell_\pol \val| ~\Bigr) \\
&\le \frac{2}{1-\disc} \min_{v\in\rep} \max_{u\in\frep(\pol)} ~\Bigl(~ u\vert_\pol\tr |\val - \Bell \val| ~\Bigr)~,
\end{align*}
where $\pol$ is a policy greedy with respect to $\val$ when not specified otherwise. Finally, the bound is established by relaxing the constraints on $u$:
\begin{align*}
\return\opt - \return(\bar\pol) &\stackrel{\text{L \ref{lem:fixed_sum}}}{\le} \frac{2}{1-\disc} \min_{v\in\rep} \max_u  ~\Bigl\{~u\vert_\pol\tr |\val - \Bell \val| \setsep \one\tr u = 1, \zero \le u\le \pol ~\Bigr\} \\
&\le \frac{2}{1-\disc} \min_{v\in\rep} \max_{u}  ~\Bigl\{~  u\vert_\pol\tr |\val - \Bell \val| \setsep \one\tr u \le 1, u\ge \zero~\Bigr\} \\
&= \frac{2}{1-\disc} \min_{v\in\rep} \max_{u}  ~\Bigl\{~ u\vert_\pol\tr |\val - \Bell \val| \setsep \| u \|_1 \le 1~\Bigr\} \\
&\stackrel{(\star)}{\le} \frac{2}{1-\disc} \min_{v\in\rep} \left\| \val - \Bell \val \right\|_\infty ~.
\end{align*}
The inequality $(\star)$ follows by the Holder's inequality.
\end{proof}
}

\cref{thm:simplelossbounds} highlights several advantages of the DRADP bound \eqref{eq:dradp_bound_simple} over \eqref{eq:api_bound}: 1) it bounds the expected loss $\|\val\opt - \val_{\bar\pol}\|_{1,\indist}$ instead of the worst-case loss $\|\val\opt - \val_{\bar\pol}\|_{\infty}$, 2) it is smaller  by a factor of $1/(1-\disc)$, 3) it holds in finite time instead of a limit, and 4) its right-hand side is with respect to the best approximation of the optimal value function instead of the worst case approximation over all iteration. In comparison with approximate linear programming bounds, \eqref{eq:dradp_bound_simple} bounds the true policy loss and not simply the approximation of $\val\opt$~\cite{Farias2003}. The limitation of \eqref{eq:dradp_bound_simple}, however, is that it relies on an $L_\infty$ norm which can be quite conservative. We address this issue in two ways. First, we  prove a bound of a different type.
\begin{thm}\label{thm:loss_bounds_direct}
Suppose that $\tilde\policies \supseteq \greedy$ and that $\bar\pol \in\arg\max_{\pol\in\tilde\policies} \tilde\return(\pol)$ in \eqref{eq:main_optimization}. Then, the policy loss $\return\opt - \return(\bar\pol)$ is bounded as:
\begin{equation} \label{eq:bound_new_type}
\|\val\opt - \val_{\bar\pol}\|_{1,\indist} \leq \min_{\val \in \rep, \val \leq \val\opt } \| \val - \val\opt \|_{1,\indist}~.
\end{equation}
\end{thm}
\choice{}{
\begin{proof}[Proof of \cref{thm:loss_bounds_direct}]
Assume that $\bar\pol$ is an optimal solution of \eqref{eq:main_optimization}. Then:\begin{align*}
\return\opt - \return(\bar\pol) &\stackrel{\text{T \ref{thm:mainloss}}}{\le}  \return\opt - \tilde\return(\bar\pol) \le \min_{\pol\in\policies_R} ~(~\return\opt - \tilde\return(\pol)~\bigr) \le \return\opt - \tilde\return(\pol\opt)~.
\end{align*}
Now, using the dual representation of $\tilde\return(\cdot)$ as in  \cref{thm:representation} (see also \cref{thm:bilinear_correct} below) we get:
\[
\return\opt - \return(\bar\pol) \le  \min \left\{ \return\opt - \indist\tr\repm\lambda_1 - \lambda_2\tr \pol\opt \setsep (1-\disc) \cdot\lambda_2 \ge \tmat \repm \lambda_1 - \trew,~ \lambda_2 \geq \zero \right\}~.
\]
It can be readily shown by subtracting $\one$ from $\repm \lambda_1$ that there exists an optimal $\lambda_1, \lambda_2$  such that $\lambda_2\tr \pol\opt = 0$ (see (ii) of \cref{thm:bilinear_correct}). Then:
\begin{align*}
\return\opt - \return(\bar\pol) &\le  \min \left\{ \return\opt - \indist\tr\repm\lambda_1   \setsep (1-\disc) \cdot \lambda_2 \ge \tmat \repm \lambda_1 - \trew,~ \lambda_2 \geq \zero,~ \lambda_2\tr \pol\opt = 0\right\} \\
&= \min \left\{ \return\opt - \indist\tr\repm\lambda_1   \setsep (\eye - \disc \cdot\tran_{\pol\opt}) \repm \lambda_1 \le \rew_{\pol\opt} \right\} \\
&= \min \left\{ \return\opt - \indist\tr\val   \setsep (\eye - \disc \cdot\tran_{\pol\opt}) \val \le \rew_{\pol\opt} , ~\val \in\rep\right\} \\
&\stackrel{\text{L \ref{lem:Bellman_monotonous}}}{=} \min \left\{ \return\opt - \indist\tr\val   \setsep \val \le \invm{\eye - \disc \cdot\tran_{\pol\opt}} \rew_{\pol\opt} , ~\val \in\rep\right\} \\
&\stackrel{\text{P \ref{prop:return_frequencies}}}{=} \min \left\{ \indist\tr
\val\opt - \indist\tr\val  \setsep \val \le \val\opt , ~\val \in\rep\right\}~.
\end{align*}
\end{proof}
}

The bound \eqref{eq:bound_new_type}, unlike bounds in most ADP algorithms, does not contain a factor of $1/(1-\disc)$ of any power. Although \eqref{eq:bound_new_type} does not involve an $L_\infty$ norm, it does require that $v\le v\opt$ which may be undesirable. Next, we show bounds that rely purely on weighted norms under additional assumptions on the concentration coefficient.

As mentioned above, \cref{asm:smooth} can be used to improve the solutions of DRADP and to derive tighter bounds. Note that this assumption must be known in advance and cannot be gleaned from the samples.  To this end, for some fixed $C\in\RealPlus$ and $\mu\in\Real^\states$ in \cref{asm:smooth}, define:
\begin{gather*}
\frep_S(\pol) = \left\{ u \in \frep(\pol) \setsep \sum_{a\in\actions} u(s,a) \le \sigma(s), ~ \forall s\in\states\right\}, \\
\sigma(s) = \disc \cdot\mu(s) + (1-\disc)\cdot \indist(s), \\
\tilde\return_S(\pol) =\max_{\val\in\rep} \min_{u \in \frep_S(\pol)} ~\Bigl(~\indist\tr \val - \frac{u\tr (\tmat \val - \trew)}{1-\disc} ~\Bigr)~.
\end{gather*}
These assumptions imply the following structure of all admissible state frequencies.
\begin{lem} \label{lem:smooth}
Suppose that \cref{asm:smooth} holds with constants $C$ and $\mu$. Then: $d \le C\cdot\sigma$ for each $d\in\frep_S(\pol)$ and $\pol\in\policies_R$.
\end{lem}
\choice{}{
\begin{proof}[Proof of \cref{lem:smooth}]
Suppose any $\pol\in\policies_R$ and its occupancy frequency $u$. Then from the definition of $u$:
\begin{align*}
d(s) &= \sum_{a\in\actions} u(s,a) = \disc\cdot  \sum_{\substack{s'\in\states\\ a'\in\actions}} \tran(s',a',s)\cdot u(s',a') + (1-\disc)\cdot \indist(s) \\
&\stackrel{u\ge \zero}{\le} \disc\cdot  \sum_{\substack{s'\in\states \\ a'\in\actions}} C\cdot \mu(s)\cdot u(s',a') + (1-\disc)\cdot \indist(s) \\
&\le \disc\cdot  C \cdot \mu(s) \Bigl(\sum_{s'\in\states,a'\in\actions}  u(s',a') \Bigr) + (1-\disc)\cdot \indist(s) \\
&\stackrel{\text{P \ref{prop:distrib_set}}}{\le} \disc \cdot C \cdot \mu(s) + (1-\disc)\cdot  \indist(s) \le C \cdot\sigma(s)~,
\end{align*}
which implies that $\frep_S(\pol) \supseteq\fmap(\pol)$ for any $\pol\in\policies_R$ . Therefore $\tilde\return_S(\pol) \le \return(\pol)$ from \cref{thm:mainloss}.
\end{proof}}

The following theorem shows a tighter bound on the DRADP policy loss for MDPs that satisfy the smoothness assumption.
\begin{thm}\label{thm:smooth}
Suppose that \cref{asm:smooth} holds with constants $C$ and $\mu$, $\tilde\policies \supseteq \greedy$, and that $\bar\pol \in\arg\max_{\pol\in\tilde\policies} \tilde\return(\pol)$ in \eqref{eq:main_optimization}. Then, the loss of $\bar\pol$ is bounded as:
\begin{equation} \label{eq:dradp_bound_smooth}
\return\opt - \return(\bar\pol) \leq \frac{2 \cdot C}{1-\disc} \min_{v\in\rep}   \left\| \val - \Bell \val \right\|_{1,\sigma} ~.
\end{equation}
\end{thm}
\choice{}{
\begin{proof}[Proof of \cref{thm:smooth}]
Using the same derivation as in the proof of \cref{thm:simplelossbounds}, we have that:
\begin{align*}
\return\opt - \return(\bar\pol) &\le \frac{2}{1-\disc} \min_{v\in\rep} \max_u  ~\Bigl\{~u\vert_\pol\tr |\val - \Bell \val| \setsep u\in\frep_S(\pol) ~\Bigr\} \\
\end{align*}
Then, the bound can be derived from the assumption as follows:
\begin{align*}
\return\opt - \return(\bar\pol) &\stackrel{\text{L \ref{lem:smooth}}}{\le} \frac{2}{1-\disc} \min_{v\in\rep} \max_u ~\Bigl\{~u\vert_\pol\tr |\val - \Bell \val| \setsep \zero \le u, ~ \sum_{a\in\actions} u(s,a) \le C \cdot \sigma(s) ~\Bigr\} \\
&\le \frac{2 \cdot C}{1-\disc} \min_{v\in\rep} \max_u ~\Bigl\{~ u\vert_\pol\tr \operatorname{diag}(\sigma) |\val - \Bell \val| \setsep  \zero\le u,~ \sum_{a\in\actions} u(s,a) \le \one \Bigr\} \\
&\le \frac{2 \cdot C}{1-\disc} \min_{v\in\rep} \max_d ~\Bigl\{~ d\tr \operatorname{diag}(\sigma) |\val - \Bell \val| \setsep  \| d \|_\infty \le 1 \Bigr\} \\
&\stackrel{(\star)}{\le} \frac{2 \cdot C}{1-\disc} \min_{v\in\rep} \|\val - \Bell \val\|_{1,\sigma} .
\end{align*}
The inequality $(\star)$ follows by Holder's inequality.
\end{proof}}

The bound in \cref{thm:smooth} is similar to comparable $L_p$ bounds for API~\cite{Munos2003}, except it relies on a weighted $L_1$ norm instead of the $L_2$ norm and preserves all the advantages of \cref{thm:simplelossbounds}. \cref{thm:smooth} exploits that the set of  occupancy frequencies is restricted under the smoothness assumption which leads to a tighter lower bound $\tilde\return_S$ on the return.

Finally, DRADP is closely related to robust ABP~\citep{Petrik2009e,Petrik2011} but provides several significant advantages. First, DRADP does not require transitive feasible~\cite{Petrik2011} value functions, which simplifies the use of constraint generation. Second, ABP minimizes $L_\infty$ bounds $\return_r : \policies_R \rightarrow \Real$ on the policy loss, which can be too conservative. In fact, it is easy to show that DRADP solutions can be better than ABP solutions by an arbitrarily large factor.

\section{Computational Models} \label{sec:computational}

In this section, we describe how to solve the DRADP optimization problem. Since DRADP generalizes ABP~\cite{Petrik2009e}, it is necessarily NP complete to solve in theory, but relatively easy to solve in practice. Note that the NP-completeness is in terms of the number of samples and features and not in the number of states or actions of the MDP. In addition, the NP completeness a is favorable property when compared to API algorithms, such as LSPI, which may never converge~\cite{Lagoudakis2003}.

To solve DRADPs in practice, we derive bilinear and mixed integer linear program formulations for which many powerful solvers have been developed. These formulations lead to anytime solvers---even approximate solutions result in valid policies---and can therefore easily trade off solution quality with time complexity.

To derive bilinear formulations of DRADP, we  represent the set of policies $\tilde\policies$ using linear equalities as:
$ \tilde\policies = \left\{ \pol\in [0,1]^\stateactions \setsep  \sum_{a \in \actions} \pol(s,a) = 1\right\}.$ This set can be defined using matrix notation as $B \pol = \one$ and $\pol \geq \zero$, where $B: |\states| \times |\stateactions|$ is defined as: $B(s',(s',a)) = 1$ when $s = s'$ and 0 otherwise. Clearly $\tilde\policies\supseteq\greedy$, which implies that the computed policy is greedy with respect to a representable value function from \cref{thm:representation} even as sampled. It would be easy to restrict the set $\tilde\policies$ by assuming the same action must be taken in a subset of states: one would add constraints $\pol(s,a) = \pol(s',a)$ for some $s,s'\in\states$ and all $a\in\actions$.

When the set of approximate policies is represented by linear inequalities, the DRADP optimization problem can be formulated as the following \emph{separable bilinear program}~\cite{Horst1996}.
\begin{mprog} \label{mpr:main_optimization_bilinear}
\maximize{\pol,\lambda_1, \lambda_2} \indist\tr\repm\lambda_1 - \pol\tr \lambda_2
\stc B \pol = \one, \qquad \pol \ge \zero, \qquad \lambda_2 \geq \zero,
\cs (1-\disc)\cdot  \lambda_2 \ge \tmat \repm \lambda_1 - \trew~.
\end{mprog}
Bilinear programs are a generalization of linear programs and are in general NP hard to solve.
\begin{thm}\label{thm:bilinear_correct}
Suppose that $\tilde\policies \supseteq \greedy$. Then the sets of optimal solutions of \eqref{mpr:main_optimization_bilinear} and \eqref{eq:main_optimization} are identical and
there exists an optimal solution $(\bar\pol,\bar\lambda_1,\bar\lambda_2)$ of \eqref{mpr:main_optimization_bilinear} such that:
\begin{enumerate}
\item[(i)] $\bar\pol$ is deterministic and greedy with respect to $\repm \bar\lambda_1$,
\item[(ii)] $\bar\pol\tr \bar\lambda_2 = 0$.
\end{enumerate}
\end{thm}
\choice{}{
\begin{proof}[Proof of \cref{thm:bilinear_correct}]
The equivalence of the sets of optimal solutions follows directly from the dual representation from the proof of \cref{thm:representation}. We next prove the two properties separately:
\begin{itemize}
\item[(i)] Follows directly from (ii) of \cref{thm:representation}.
\item[(ii)] To prove by contradiction that there is an optimal solution $\pol$ with $\pol\tr \lambda_2 = 0$, assume an optimal solution $\pol$ such that $\pol\tr\lambda_2 > 0$. Since  $\pol \ge\zero, ~\lambda_2\ge \zero$,  there exists an index $i$ (when treating $\pol$ as a simple vector), such that $\pol(i) = 1$ and $\lambda_2(i) > 0$ (from the integrality of $\pol$). We then show that there exists a solution with a lower value $\pol\tr \lambda_2$, smaller number of $j$ such that $\lambda_2(j) > 0$, without decreasing the objective value.
From the optimality of $\lambda_2$ and the positivity of $\lambda_2(i)$, we have that (the constraint in \eqref{mpr:main_optimization_bilinear} is active):
\[ (1-\disc)\cdot \lambda_2(i) = a_i\tr \repm \lambda_1 - b_i > 0, \]
where $a_i\tr$ is the $i$-th row of the matrix $\tmat$ and $b_i$ is the $i$-th element of the vector $\trew$.  Now, from \cref{asm:contains_one}, there exists $\lambda_1'$ such that:
\[ \repm \lambda_1' = \repm \lambda_1 - \lambda_2(i)\cdot \one. \]
This $\lambda_1$ is trivially feasible in \eqref{mpr:main_optimization_bilinear}. Let $(1-\disc)\cdot \lambda_2' = \pos{\tmat \repm \lambda_1' - \trew}$. Now using  \cref{lem:Bellman_one} it is readily seen that: $\tmat\, \one = (1-\disc)\cdot \one$. Then:
\begin{align*}
\tmat \repm \lambda_1' - \trew &= \tmat \repm \lambda_1 - \trew -  \lambda_2(i)\cdot \one, \\
\tmat \repm \lambda_1' - \trew &\le \tmat \repm \lambda_1 - \trew, \\
\pos{\tmat \repm \lambda_1' - \trew} &\le \pos{\tmat \repm \lambda_1 - \trew}, \\
\lambda_2' &\le \lambda_2.
\end{align*}
In addition, $\lambda_2'(i) = 0$, and thus: $\lambda_2' \le \lambda_2 - \lambda_2(i)$. Then:
\begin{align*}
\indist\tr \repm \lambda_1' - \pol\tr \lambda_2' &\ge \indist\tr \repm \lambda_1 - \lambda_2(i) - \pol\tr \lambda_2 + \lambda_2(i)
\ge \indist\tr \repm \lambda_1 - \pol\tr \lambda_2~.
\end{align*}
Therefore $\lambda_1',\lambda_2'$ are also optimal with one less nonzero index $i$ such that $\pol(i) > 0$ and $\lambda'_2(\pol)>0$. The claim then follows by choosing $i$ with the maximal $\lambda_2(i)$.
\end{itemize}
\end{proof}}

Because there are few, if any, industrial solvers for bilinear programs, we reformulate \eqref{mpr:main_optimization_bilinear} as a mixed integer linear program~(MILP). Any separable bilinear program can be formulated as a MILP~\citep{Horst1996}, but such generic formulations are impractical because they lead to large MIL{Ps with weak linear relaxations. Instead, we derive a more compact and structured MILP formulation that exploits the existence of optimal deterministic policies in DRADP (see (i) of \cref{thm:bilinear_correct}) and is based on McCormic inequalities on the bilinear terms~\cite{Linderoth2005}. To formulate the MILP, assume a given upper bound $\tau\in\Real$ on any optimal solution $\lambda_2\opt$ of \eqref{mpr:main_optimization_bilinear} such that $\tau \geq \lambda_2\opt(s,a)$ for all $s\in\states$ and $a\in\actions$. Then:
\begin{mprog} \label{mpr:dradp_milp}
\maximize{z,\pol,\lambda_1,\lambda_2} \indist\tr \repm \lambda_1 -  \one\tr z
\stc z \geq \lambda_2 - \tau\cdot(\one - \pol),
\cs (1-\disc) \cdot\lambda_2 \geq \tmat \repm \lambda_1 - \trew,
\cs B \pol = \one, \quad \pol \in \{0,1\}^{|\states| |\actions|}
\cs z \ge \zero, \quad \lambda_2 \geq \zero ~.
\end{mprog}

\begin{thm}\label{thm:dradp_milp}
Suppose that $\tilde\policies \supseteq \greedy$ and $(\bar\pol,\bar\lambda_1, \bar\lambda_2, \bar{z})$ is an optimal solution of \eqref{mpr:dradp_milp}. Then, $(\bar\pol, \bar\lambda_1,\bar\lambda_2)$ is an optimal solution of \eqref{mpr:main_optimization_bilinear} with the same objective value given that  $\tau > \| \bar\lambda_2\|_\infty$.
\end{thm}
\choice{}{
\begin{proof}[Proof of \cref{thm:dradp_milp}]
We use McCormic inequalities, as described in \cite{Linderoth2005}. In particular, assume that $(x,y) \in \{ (x,y) \setsep l_x \le x \le u_x, l_y \le  y \le u_y\}$. Then the following holds:
\begin{align*}
x\cdot y &\ge \max\{ l_y\cdot x + l_x\cdot y - l_x\cdot l_y , u_y\cdot x + u_x\cdot y - u_x\cdot u_y \} \\
x\cdot y &\le \min\{ u_y\cdot x + l_x\cdot y - l_x\cdot u_y , l_y\cdot x + u_x\cdot y - u_x\cdot l_y \}~.
\end{align*}
To simplify the notation, we use a single index $i \in \stateactions$ to denote the individual elements of $\pol, \lambda_2, z$. From the assumptions of the theorem and the constraints on policies, we have that: $\pol(i) \in [0,1]$ and $\lambda_2 \in [0,\tau]$. Then for all $i$ the McCormic inequalities for \eqref{mpr:main_optimization_bilinear} are:
\begin{align*}
\pol(i)\cdot \lambda_2(i) &\ge \max \{ 0, \lambda_2(i) - \tau\cdot (1 - \pol(i)) \} \\
\pol(i)\cdot \lambda_2(i) &\le \min\{ \tau\cdot \pol(i), \lambda_2(i) \}.
\end{align*}

Since any optimal $z$ in \eqref{mpr:dradp_milp} satisfies $z(i) = \max \{ 0, \lambda_2(i) - \tau\cdot (1 - \pol(i)) \}$, $z(i)$ is a \emph{lower} bound on $\pol(i)\cdot \lambda(i)$. As a result, for any $\pol\in\policies_R$, the objective of \eqref{mpr:dradp_milp} is an \emph{upper} bound on the objective of \eqref{mpr:main_optimization_bilinear}.  Then, it can be readily shown that whenever $\pol(i) \in \{0,1\}$ then $\pol(i) = \lambda_2(i) \pol(i)$: when $\pol(i) = 1$ then $z(i) = \lambda_2(i) = \pol(i) \cdot\lambda_2(i)$ and when $\pol(i) = 0$ then $z(i) = \max\{0, \lambda_2(i) - \tau \cdot(1 - \pol(i)) \} = 0 = \pol(i) \cdot\lambda_2(i)$. This shows the equality of the sets of optimal solutions stated in the theorem.
\end{proof}}

As discussed above, any practical implementation of DRADP must be sample-based. The bilinear program \eqref{mpr:main_optimization_bilinear} is constructed from samples very similarly to ALPs (e.g. Sec 6 of \cite{Farias2003}) and identically to ABPs (e.g. Sec 6 of \cite{Petrik2011}). Briefly, the formulation involves only the rows of $\tmat$ that correspond to transitions of sampled state-action pairs and $\trew$ entries are estimated from the corresponding rewards. As a result, there is one $\lambda_1$ variable for each feature, and $\lambda_2$ and $\pol$ are nonzero only for the sampled rows of $\tmat$ (zeros do not need to be considered). The size of the optimization problem \eqref{mpr:main_optimization_bilinear} is then independent of the number of states and actions of the MDP; it depends only on the number of samples and features.

\section{Experimental Results} \label{sec:experimental}

In this section, we experimentally evaluate the empirical performance of DRADP. We present results on the inverted pendulum problem---a standard benchmark problem---and a  synthetic chain problem. We gather state and action samples in advance and solve MILP \eqref{mpr:dradp_milp} using IBM CPLEX 12.2. We then compare the results to three related methods which work on offline samples: 1) LSPI~\cite{Lagoudakis2003}, 2) ALP~\cite{Farias2003}, and 3) ABP~\cite{Petrik2009e}.
While  solving the MILP formulation of DRADP is NP hard (in the number of features and samples), this does not mean that the computation takes longer than for other ADP methods; for example, the computation time of LSPI is unbounded in the worst case (there are no convergence guarantees). In the experiments, we restrict the computation time for all methods to 60s.

\paragraph{Inverted Pendulum} The goal in the inverted pendulum benchmark problem is to balance an inverted pole by accelerating a cart in either of two directions~\citep{Lagoudakis2003}. There are three actions that represent applying the force of $u = -50 N$, $u = 0 N$, and $u = 50 N$ to the cart with a uniform noise between $-10 N$ and $10 N$. The angle of the inverted pendulum is governed by a differential equation. We used the standard features for this benchmark problem for all the methods: 9 radial basis functions arranged in a grid over the 2-dimensional state space with centers $\mu_i$ and a constant term required by \cref{asm:contains_one}. The problem setting, including the initial distribution is identical to the setting in \cite{Lagoudakis2003}.

\begin{figure}
\centering
\includegraphics[width=0.45\textwidth]{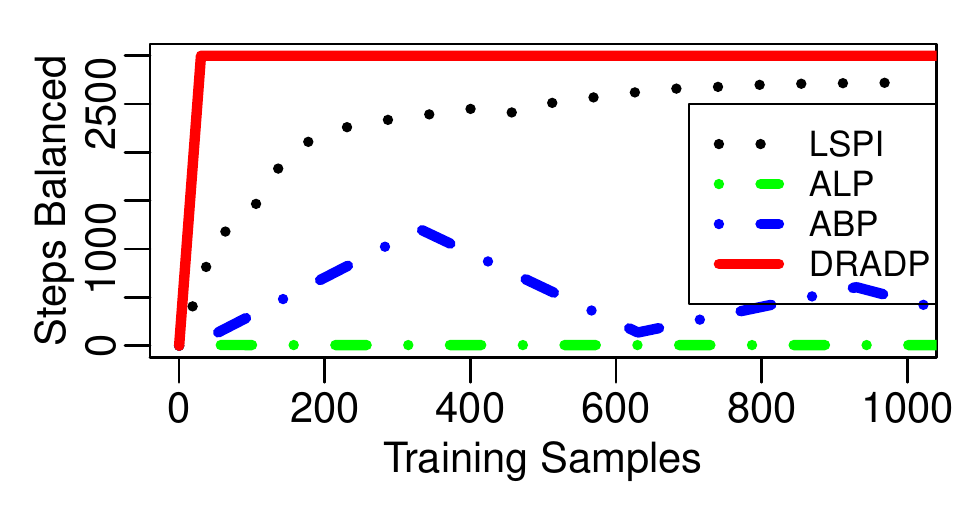}
\caption{Inverted pendulum results} \label{fig:inverted_pendulum}
\end{figure}

\begin{figure}
\centering
\includegraphics[width=0.45\textwidth]{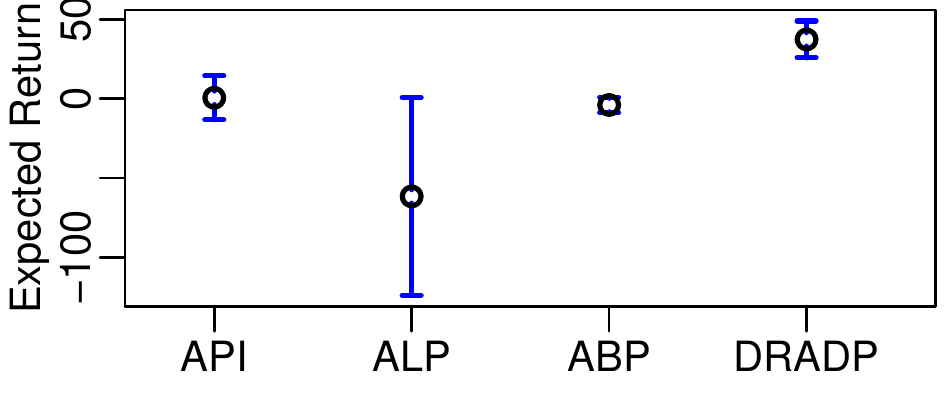}
\caption{Expected return on the chain benchmark} \label{fig:chain}
\end{figure}

\cref{fig:inverted_pendulum} shows the number of balancing steps (with a 3000-step bound) for each method as a function of the number of training samples averaged over 5 runs. The figure does not show error bars for clarity; the variance was close to 0 for DRADP. The results indicate that DRADP computes a very good solution for even a small number of samples and significantly outperforms LSPI. Note the poor performance of ABP and ALP with the 10 standard features; better results have been obtained with large and different feature spaces~\citep{Petrik2010} but even these do not match DRADP. The solution quality of ABP decreases with more samples, because the bounds become more conservative and the optimization problems become harder to solve.

\paragraph{Chain Problem}

Because the solution quality of ADP methods depends on many factors, good results on a single benchmark problem do not necessarily generalize to other domains. We, therefore, compare DRADP to other methods on a large number of randomly generated chain problems. This problem consists of 30 states $s_1 \ldots s_{30}$ and 2 actions: left and right with 10\% chance of moving the opposite way. The features are 10 orthogonal polynomials. The rewards are 0 except:  $\rew(s_2) = -50, ~\rew(s_3) = 4, ~\rew(s_4) = -50, ~\rew(s_{20}) = 10$. \cref{fig:chain} shows the results of 1000 instantiations with randomly chosen initial distributions and indicates that DRADP significantly outperforms other methods including API (a simple version of LSPI).

\section{Conclusion}

This paper proposes and analyzes DRADP---a new ADP method. DRADP is based on a mathematical optimization formulation---like ALP---but offers significantly stronger theoretical guarantees and better empirical performance. The DRADP framework also makes it easy to improve the solution quality by incorporating additional assumptions on state occupation frequencies, such as the small concentration coefficient. Given the encouraging theoretical and empirical properties of DRADP, we hope it will lead to better methods for solving large MDPs and will help to deepen the understanding of ADP.

\paragraph*{Acknowledgements}

I thank Dan Iancu and Dharmashankar Subramanian for the discussions that inspired this paper. I also thank the anonymous ICML 2012 and EWRL 2012 reviewers for their detailed comments.

\begin{small}
\bibliographystyle{icml2012}
\bibliography{../../MainReferences}
\end{small}

\choice{}{
\clearpage
\appendix
\onecolumn

\section{Basic Properties of Value Functions}

\begin{lem} \label{lem:Bellman_one}
For any  $\val\in\values$ the following holds:
$\Bell (\val + k \one) =   \Bell \val + \disc k \one.$ In addition, the sets of greedy policies with respect to $\val$ and $\val+k\one$ are identical.
\end{lem}

\begin{lem} \label{lem:Bellman_monotonous}
The operators $P$ and $(\eye - \disc P)^{-1}$ are monotonous for any stochastic matrix $P$:
\begin{align*}
x \geq y &\Rightarrow P x \geq P y \\
x \geq y &\Rightarrow (\eye - \disc P)^{-1} x \geq (\eye - \disc P)^{-1} y
\end{align*}
for all $x$ and $y$.
\end{lem}

\begin{lem}
Suppose that $\val\in\values$ satisfies $\val\ge \Bell\val$. Then $\val\ge\val\opt$.
\end{lem}

\begin{lem}
Each $\val\in\values$ satisfies: $\val-\Bell\val \le \val - \Bell_\pol \val$.
\end{lem}
}

\end{document}